\documentclass[runningheads]{llncs}
\usepackage[T1]{fontenc}

\usepackage{graphicx}
\usepackage{tabularx}
\usepackage{placeins}
\usepackage{float}
\usepackage{marvosym}

\usepackage{booktabs} 
\usepackage{xcolor} 
\usepackage{colortbl} 

\usepackage{color}
\definecolor{myred}{rgb}{.8,.0,.0}

\definecolor{mygreen}{rgb}{0.0, 0.6, 0.0}

\usepackage{booktabs}
\usepackage{caption}
\usepackage{subcaption}
\usepackage{framed,multirow}
\usepackage{amsmath}

\usepackage[breaklinks=true,bookmarks=false]{hyperref}

\begin{document}

\title{Robustness and sex differences in skin cancer detection: logistic regression vs CNNs}

\titlerunning{Robustness and sex differences in skin cancer detection}

\author{Nikolette Pedersen\inst{1} \and
Regitze Sydendal\inst{1}\textsuperscript{(\Letter)} \and
Andreas Wulff\inst{1} \and  \\
Ralf Raumanns\inst{1,3,4} \and 
Eike Petersen\inst{2} \and
Veronika Cheplygina\inst{1}
}

\authorrunning{N. Pedersen et al.}

\institute{IT University of Copenhagen, Denmark\\
\email{\{nizp,resy,lawu,ralr,vech\}@itu.dk} 
\and
Fraunhofer Institute for Digital Medicine MEVIS, Germany
\email{eike.petersen@mevis.fraunhofer.de}
\and
Fontys University of Applied Science, Eindhoven, The Netherlands
\and
Eindhoven University of Technology, Eindhoven, The Netherlands}

\maketitle 

\begin{abstract}

Deep learning has been reported to achieve high performances in the detection of skin cancer, yet many challenges regarding the reproducibility of results and biases remain. This study is a replication (different data, same analysis) of a previous study on Alzheimer's disease detection, which studied the robustness of logistic regression (LR) and convolutional neural networks (CNN) across patient sexes. We explore sex bias in skin cancer detection, using the PAD-UFES-20 dataset with LR trained on handcrafted features reflecting dermatological guidelines (ABCDE and the 7-point checklist), and a pre-trained ResNet-50 model. We evaluate these models in alignment with the replicated study: across multiple training datasets with varied sex composition to determine their robustness. Our results show that both the LR and the CNN were robust to the sex distribution, but the results also revealed that the CNN had a significantly higher accuracy (ACC) and area under the receiver operating characteristics (AUROC) for male patients compared to female patients. 
The data and relevant scripts to reproduce our results are publicly available (\url{https://github.com/nikodice4/Skin-cancer-detection-sex-bias}).

\end{abstract}

\section{Introduction} \label{sec:intro}
Classification of skin lesions using machine learning (ML), especially deep learning (DL), has shown promising results \cite{pham_ai_2021,Brinker2019}. However, such models have also been shown to suffer from spurious correlations or shortcut learning~\cite{winkler2019association,banerjee2023shortcuts,JimenezSanchez2025} as well as biases across patient sex~\cite{abbasi2020risk,larrazabal2020gender} or skin types~\cite{Daneshjou2022}.

Bias in medical imaging is an active topic of investigation.
Especially in the context of DL, it is challenging to discern what the classifier is learning, rendering bias investigations challenging. For example, Gichoya et al.~\cite{gichoya2022ai} show that DL models can reliably predict sensitive demographic variables from chest x-rays, even when these variables are not recognizable to clinical experts. While DL models are highly popular, prior evidence suggests that at least in some applications, simpler models may be more robust to shortcuts and biases while obtaining similar performance~\cite{Petersen2022}.

We investigate sex-related performance and robustness in skin lesion classification with two models: (i) a baseline LR with handcrafted features such as asymmetry, border, and colour, and (ii) a CNN trained on raw images of skin lesions. We aim to replicate (\emph{not} reproduce, since we use different data) a previous study on MRI-based Alzheimer's classification \cite{Petersen2022} which examined the robustness of LR vs CNNs, and found that LR is more robust to different dataset compositions, while CNNs (surprisingly) generally improved their performance for both male and female patients when including more female patients in the training dataset. Our experiments on skin lesion classification with the PAD-UFES-20 dataset \cite{pacheco2020pad} reach similar conclusions. Our contributions are as follows:
\begin{enumerate}
\item We replicate the results of a robustness study on Alzheimer's classification across sexes in a different medical domain.
\item We highlight previously unreported errors (such as identical lesion IDs with different patients) in the PAD-UFES-20 dataset, demonstrating the importance of data exploration before training.
\item We show that traditional methods such as handcrafted features with LR are still worth examining in the context of skin lesion classification, among others due to their robustness to biases and interpretability.
\end{enumerate}

\section{Related Work} \label{sec:related}
There are various dermatological guidelines for diagnosing lesions, including the ABCDE method and the 7-point checklist \cite{Abbasi2004,dermlite_7-point_nodate}. Early work on automatic diagnosis of skin lesions often focused on implementing ``handcrafted'' features to reflect these guidelines and applying ML techniques such as random forests. Later, DL became the dominant approach for skin cancer detection, often with CNNs, and more recently, transformers. Promising results have been reported, for example, on the International Skin Imaging Collaboration (ISIC) 2019 data  where Pham et al.~\cite{pham_ai_2021} report to have achieved performances surpassing dermatologists. 

However, many methods have been shown to suffer from biases and shortcuts. In skin lesion classification, biases have been shown for age, sex, and Fitzpatrick skin types \cite{abbasi2020risk,groh_evaluating_2021}, while shortcuts can be introduced by surgical markers \cite{winkler2019association} or other factors such as the type of dermatoscope used, lightning conditions, the presence of hair, skin tone and others. A categorisation of shortcuts and their influence on skin lesion classification is provided by Jiménez-Sánchez et al.~\cite{JimenezSanchez2025}. 

Findings on sex bias in skin lesion classification diverge, possibly due to differences in the datasets used. Some works \cite{abbasi2020risk} show performance disparities, while others no significant differences between male and female patients \cite{sies_does_2022,raumanns2024dataset}. However, Sies et al.~\cite{sies_does_2022} highlight that sex bias cannot be ruled out, especially when considering CNNs.

Several studies highlight the problems with publicly available datasets used for ML, such as underrepresentation of demographic groups or the absence of metadata needed to investigate such biases~\cite{daneshjou2021lack,wen2022characteristics}. A popular source of data is the large ISIC archive, though different studies use different subsets of it. However, it contains duplicates, and metadata identifying these is missing~\cite{cassidy2022analysis}. By contrast, the PAD-UFES-20 dataset~\cite{pacheco2020pad} with over 2K images has a more diverse representation and rich metadata. Another recent and diverse dataset is DDI~\cite{Daneshjou2022} which, however, contains less than 1000 images.

\section{Methods} \label{sec:methods}
\subsection{Data}
\begin{table}[tb]
\centering
\captionsetup{justification=centering}
\caption{Summary table of data characteristics}
\label{tab:data}
\vspace{.5em}
\begin{tabular}{lccr}
\toprule
\textbf{Lesion type} & \textbf{No. female patients} & \textbf{No. male patients} & \textbf{Total (\%)} \\
\midrule
Non-cancerous lesions & 198 & 148 & 346 (29.4) \\
Cancerous lesions & 401 & 432 & 833 (70.6) \\
\midrule
\textbf{Total} & \textbf{599} & \textbf{580} & \textbf{1179 (100)} \\
\bottomrule
\end{tabular}
\vspace{.5em}
\end{table}
We use the PAD-UFES-20 dataset \cite{pacheco2020pad}, containing 2298 images of 1641 skin lesions from 1373 patients. There are six different diagnoses, which we grouped into non-cancerous vs. skin cancer due to the size of the dataset. We removed all patients with missing entries for the gender variable\footnote{Refer to the discussion section for details on our use of the terms `sex' and `gender'.}. Furthermore, we removed any duplicates with the same lesion ID, keeping only the first occurrence.
The result is a total of 1179 lesions, a summary of the data characteristics can be seen in Table \ref{tab:data}.

During data exploration, we observed several inconsistencies in the dataset.
These included different patients with the same lesion ID and different lesion IDs for identical images. The creators of PAD-UFES-20 confirmed to us that these were indeed mistakes, and we updated the lesion IDs accordingly\footnote{Refer to our Github repository for details.}.

Non-experts (data science students and the authors) created lesion segmentations using LabelStudio~\cite{LabelStudio}. We made the segmentations available on Zenodo \footnote{\url{https://doi.org/10.5281/zenodo.16535326}}.
We upsample the non-cancerous lesions, augmenting them by flipping the images, randomly sharpening, and adding Gaussian blur. We did not perform colour augmentation, as this could affect the ``ground truth'' diagnostic features of the images.

\subsection{Feature extraction} 
We used feature extraction methods related to the ABC criteria, leaving out diameter and evolution, and the 7-point checklist. The features were originally implemented by students in a data science project course on image analysis. We selected several methods based on the degree of quality of the code and feature explanations, and reproducibility of the code. The initial set of features included:

\begin{description}
\item[Asymmetry] rotates the lesion mask multiple times by 22.5\%, ``folds'' the lesion mask to measure the overlap at this angle, and averages the resulting scores.
\item[Compactness] is defined as $c = \frac{p^2}{4 \pi A}$, where \(p\) is the perimeter of the lesion border and \(A\) is the area of the lesion. 
\item[Mean and variance of HSV] takes the means and variances of the lesion pixels for each channel.
\item[Dominating HSV] clusters lesion pixels on their full HSV values using k-Means, and selects the hue value for the largest cluster.
\item[Colour variance] clusters lesion pixels using SLIC superpixel segmentation on their RGB values, and takes the mean RGB values for each segment. 
\item[Relative colours] is defined by \cite{celebi_automatic_2008}. It clusters pixels with SLIC and takes the relative proportions red (F1) and green (F2) inside the lesions. F11 is the difference in red between the skin and the lesion. 

\item[Blue whitish veil] is defined by \cite{8357434} and counts the number of pixels where the RGB value satisfies $R > 60, R - 46 < G < R + 15$.
\end{description}

We performed feature selection using the training-validation set by removing redundant features where the Pearson correlation with more than three other features was above 0.8. After discarding the highly correlated features, we selected (again using only the training-validation set) the 10 features with the highest correlation with the labels. The features selected following this process were \texttt{mean\_asymmetry}, \texttt{compactness\_x}, \texttt{sat\_var}, \texttt{avg\_hue}, \texttt{dom\_hue}, \texttt{avg\_green\_channel}, \texttt{F1}, \texttt{F2}, \texttt{F11} and \texttt{blue\_veil\_pixels}.

\subsection{Models}
We use two models in our experiments: LR and the ResNet-50 CNN. 
The LR model (scikit-learn \cite{scikit-learn_scikit-learn_nodate} implementation) relies on the ten selected standardised handcrafted features. We then use grid search on the training-validation set to find the best C, with inverse regularisation strength parameter $C \in \{0.01, 0.05, 0.1, 0.5, 1, 2, 5\}$.
 
For the ResNet-50, we use the torchvision \texttt{IMAGENET1K\_V2} weights, fine-tuned on PAD-UFES-20. We use the ResNet-50 because it was the best performing model in a prior study performed by the dataset authors~\cite{pacheco_impact_2020}. 
We modify the last layer to adapt the model to binary classification. Before training, we resize the images to $224 \times 224$ pixels and normalise the intensities according to standard and mean values obtained from ImageNet.
We use binary cross-entropy as the loss function and the Adam optimiser with a learning rate of 0.001. We train for 10 epochs with a batch size of 32 (shuffling enabled). Hyperparameter optimisation was not performed due to limited computational resources.

\subsection{Experiments}
\subsubsection{Data splits.} For evaluation, we adopt the setup of \cite{Petersen2022} with some modifications due to the format of our data. We first split the data into four categories: \texttt{cancer\_female}, \texttt{non\_cancer\_female}, \texttt{cancer\_male} and \texttt{non\_cancer\_male}. We then create five held-out test sets with 104 patients each, consisting of 26 patients from each category. For each test set, all remaining data (including the data from the other test sets) creates a training-validation set. There are no overlaps between each training-validation and test set pair. To avoid data leakage, we kept lesions from the same patient and any augmented lesions in the same set.  

We then sample the training-validation sets according to five ratios of female patients: $0\%$, $25\%$, $50\%$, $75\%$, $100\%$, creating a total of 25 training-validation sets for each test set, and thus 125 training-validation sets in total\footnote{\href{https://github.com/nikodice4/Skin-cancer-detection-sex-bias/blob/main/split/Flowchart_of_splits.png}{Refer to our Github repository for a figure illustrating the data splits.}}.
This procedure creates 125 training-validation sets with five held-out test sets for the LR. For the CNN, we use 50 training-validation datasets and two of the five held-out test sets, due to the limitations of our computational resources. (Note that our focus is not on comparing LR vs. CNN performance, but rather to evaluate their respective robustness.)
Finally, we evaluate the training ratio impact by comparing the performance on female vs. male patients using the area under the receiver operating characteristic (AUROC) and accuracy (ACC).

\noindent \subsubsection{Statistical significance tests.}  To test for statistically significant differences across sex ratios, we perform a regression t-test. The null hypothesis is $H_0: m=0$, where $m$ is the slope of the linear regression of the ACC or AUROC across all sex ratios. With multiple comparisons, patient sex (female/male), model type (CNN/LR), and evaluation metric (ACC/AUROC), lead to a total of eight ($2ˆ3$) tests. We therefore apply a (conservative) Bonferroni correction to avoid Type I errors, leading to a corrected threshold of $\alpha = 0.006$. We then use a Mann-Whitney U-test to test for performance differences between the sexes. 
Our significance tests are used for assessing the robustness of each model when trained on different female/male patient ratios and then tested on female/male patients. We do not aim to compare LR vs. CNN performance.

\section{Results} \label{sec:results}
We show the results for the LR and CNN models in Table \ref{tab:metrics}. Although we cannot directly compare the models, we do not observe large differences in the results, with the mean performance for LR being slightly higher than that of the CNN in half of the reported metrics. Our AUROC and ACC scores are lower than, for example, reported by Pacheco et al.~\cite{pacheco_impact_2020}, however, they also included other clinical factors in their models. 

\begin{table}[tb]
\centering
\caption{Mean and standard deviation of ACC and AUROC for LR and CNN}
\label{tab:metrics}
\captionsetup{justification=centering} 
\vspace{.5em}
\begin{tabular}{@{}p{3cm}p{2.5cm}p{2.5cm}@{}}
\toprule
\textbf{Metrics} & \textbf{LR} & \textbf{CNN} \\
\midrule
ACC, f & 0.682 ± 0.059 & 0.650 ± 0.057 \\
ACC, m & 0.684 ± 0.080 & 0.687 ± 0.053 \\ 
AUROC, f & 0.745 ± 0.052 & 0.716 ± 0.075 \\ 
AUROC, m & 0.727 ± 0.081 & 0.758 ± 0.059 \\ 



\bottomrule
\end{tabular}
\vspace{.5em}
\end{table}

\begin{figure*}[tb]
  \centering
  \includegraphics[width=0.77\linewidth]{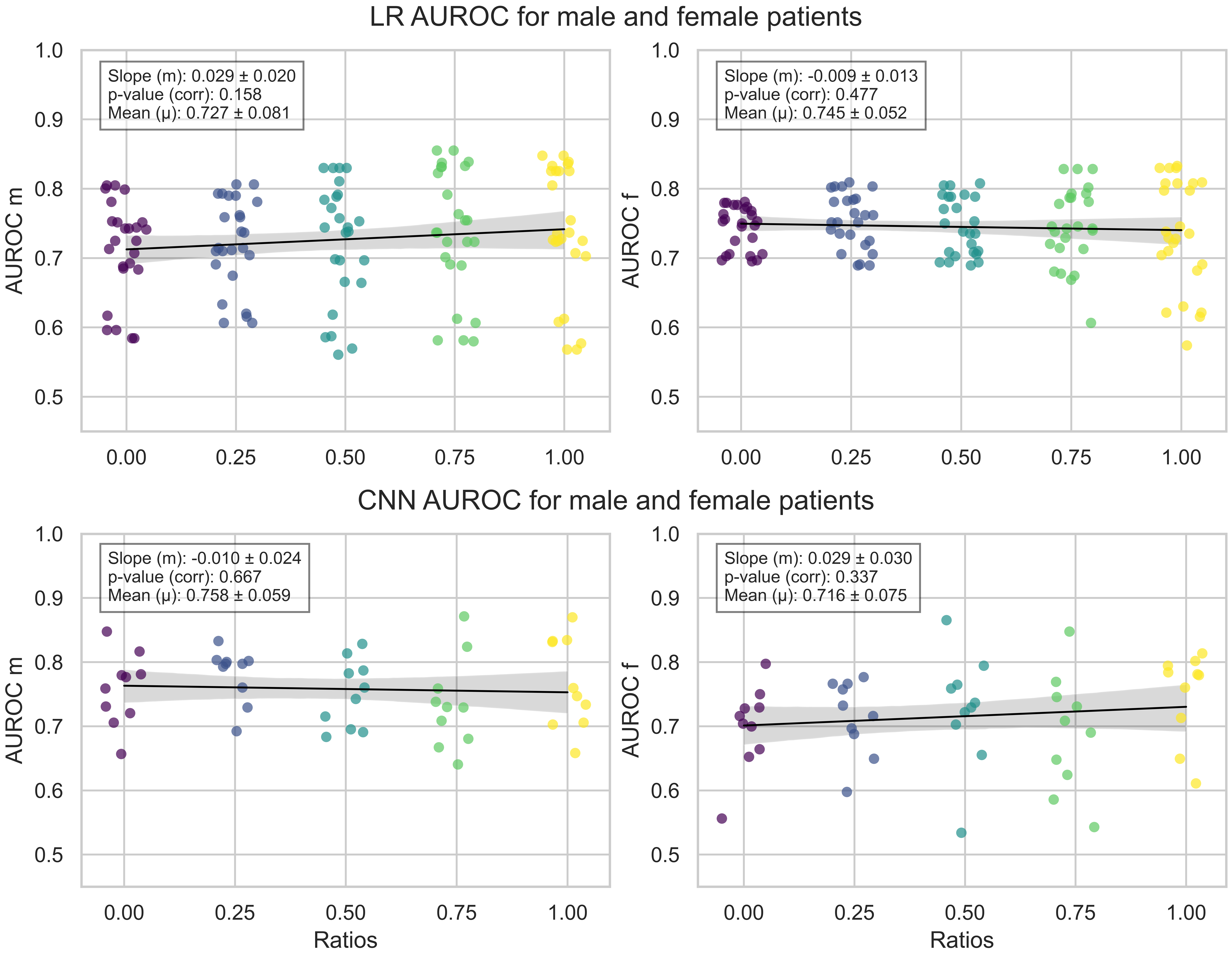} 
  \caption{AUROC evaluated on male (left) and female (right) patients, for LR (top) and CNN (bottom) across different training sex ratios, where a ratio of 1.00 corresponds to a training set entirely of female patients.}
  \label{fig:LR}
\end{figure*}

Figure \ref{fig:LR} shows the performance metrics of the LR and CNN, trained on the different sex ratios and then tested separately on female and male patients. We show the results of the statistical significance tests in Table \ref{tab:null}. None of the $p$-values for the regression t-test fall below the corrected threshold, and therefore we consider none of them statistically significant. 
\begin{table*}[tb]
\centering
\caption{The \textit{p}-values of the significance tests for LR and the CNN}
\label{tab:null}
\captionsetup{justification=centering}
\vspace{.5em}
\begin{tabular}{p{6.8cm} c r}
\toprule
\textbf{Statistical test} & Threshold & \textbf{\textit{p}-value}\\
\midrule
\textbf{Regression t-test} & $\frac{0.05}{8} = 0.006$ & \\
\midrule
LR AUROC, f &  & 0.477 \\
LR AUROC, m &  & 0.158 \\
CNN AUROC, f &  & 0.3374 \\
CNN AUROC, m &  & 0.6674 \\
\midrule
\textbf{Mann-Whitney U test} & $\frac{0.05}{4} = 0.0125$ & \\
\midrule
LR AUROC (f \& m) &  & 0.219 \\
CNN AUROC (f \& m) &  & 0.007 \\
\bottomrule
\end{tabular}

\end{table*}

We further analyse the performance variability in Figure \ref{fig:LR}. For LR, for female patients the lowest standard deviation of the AUROC is at ratio 0.0, steadily increasing until ratio 1.0. For male patients we do not observe the same pattern, as the scatter seems consistent across all ratios. In contrast, for the CNN, we see for all plots that the smallest standard deviations are at ratio 0.5.

In Table \ref{tab:null}, the Mann-Whitney U-test results show that for the CNN model, there are statistically significant differences in the AUROC between male and female patients, with male patients having higher performance. 

\section{Discussion and Conclusions} \label{sec:discussion}
\subsubsection{Summary of results.} The results show that for both LR and CNN, the null hypothesis of the models being robust to a data shift change could not be rejected. However, for the CNN, we found that the mean of the ACC and AUROC scores are statistically significantly higher for male than for female patients, which was not the case for the LR: LR is more robust. 

Even though we could not reject the null hypothesis, it is worth mentioning that the slope for female patients for the CNN's ACC scores is positive, even when accounting for uncertainty. The slope might have been different if more computational resources had been dedicated to the CNN training. However, in contrast to a previous study \cite{sies_does_2022}, we found that for the CNN, the performance was better for male patients in general. We hypothesise that one of the reasons why we found this difference and \cite{sies_does_2022} did not could be because they only processed dermoscopic images, which might have led to less difference between the sexes since the images are generally of a higher quality and the lesions are better visible than in the smartphone images of which PAD-UFES-20 consists.

\subsubsection{Limitations.}

Our results should be interpreted carefully due to the data and models we used. The PAD-UFES images are taken with smartphones, rather than with a dermatoscope used in other datasets, so our results might not generalise to studies on other datasets. We found multiple errors in the PAD-UFES-20 dataset which we then corrected, but we cannot guarantee that all errors are identified. Therefore, our conclusions might not necessarily hold if, for example, the dataset would be audited or relabeled by experts.  

Furthermore, although we split the dataset carefully to use as much data as possible, the amount of training data was relatively small. However, we believe that the rich meta-data in PAD-UFES-20 justifies using smaller sample sizes to investigate robustness, as datasets collected in similar conditions would not necessarily be large.

The handcrafted features used by the LR rely on segmentation masks, which were created manually by students without expertise in dermatology. A different segmentation method would result in different feature values. The CNN used the raw images, but it would be interesting to investigate how incorporating the masks during training would affect the performance. 

The LR and CNN models did not reach state-of-the-art performance, although this was also not our goal as we focused on robustness and fairness. In future work, it would be worth investigating other popular architectures. Our code is available and reproducible and allows for such extended experiments. 

\subsubsection{Ethical considerations.}
Biases are not just a feature of the distribution in the dataset and often cannot be disentangled from other factors. For example, people with different skin types are likely to have different incidences of skin cancer. Other causes of bias could include medical education being primarily centred around people with lighter skin \cite{balch_why_2022}, and a general lack of access to healthcare for patients belonging to racial and ethnic minorities.

We used the term sex in this paper instead of the term gender as the dataset only mentions male and female individuals, and in medical imaging datasets in general, sex is the variable that is often collected. However, in reality, sex is not binary either, and we acknowledge that simply changing the name of the variable might cause problems for individuals whose gender identity does not align with their biological sex. 

Finally, deep learning has a significant carbon footprint. While the scale of our project is comparatively small due to the use of traditional methods and a smaller CNN, we believe researchers should be considerate of the carbon footprint of using (larger and larger) models which could still have biases across different patient groups.


\subsubsection{Concluding remarks.} We investigated the robustness of a logistic regression model and a ResNet-50 CNN in the context of skin cancer detection in the PAD-UFES-20 dataset. We investigated different ratios of sex compositions in the training data. Our findings show that both classifiers were robust across the different distributions (the CNN less so, but not statistically significantly), which replicates the conclusions of a study on sex differences in Alzheimer's classification~\cite{Petersen2022} and differs from earlier results in chest x-ray diagnosis~\cite{larrazabal2020gender}.
However, overall, the CNN had consistently higher performance for male patients. Therefore, even though both models showed robustness to the dataset distribution, the CNN's better performance in male patients suggests a need for further research into the causal factors that influence model biases, both in skin cancer detection specifically and in the broader field of medical image analysis.

For future work in bias assessment and mitigation, our primary recommendation is to consider model choice as a key factor.
Simpler models with hand-crafted features are often neglected in current studies in favour of modern deep learning approaches, but both our study and the one by Petersen et al.~\cite{Petersen2022} suggest that simpler, traditional models may perform similarly well and be more robust to dataset shifts and biases.
These findings are in line with recent lines of work on model multiplicity~\cite{Black2022} and representation learning~\cite{Kumar2025}, which suggest that there are often many (superficially) similarly well-performing models with otherwise very distinct characteristics for a given task.


\begin{credits}
\subsubsection{\ackname}\label{sec:Acknowledgments}
We acknowledge the student groups at IT University of Copenhagen for their annotations: Bees (5), Cute Cats Combating Cancer (4), Dragons (5), Elephants (5), Flamingo (5), GroupIT (5), Hedgehawks (5), Iguanas (soaking the sun) (6), Jaguars (5), Koalas (5), Magical Moles (5), Okapi (5), Possum (4) and
Queen Snakes (4).

\subsubsection{\discintname}\label{sec:Interests}
The authors declare that they have no competing interests.

\end{credits}

%
%
\bibliographystyle{splncs04}
\bibliography{refs_students_f,refs_veronika_f,refs_public}

\end{document}